\documentclass[a4paper,conference]{IEEEtran}
\IEEEoverridecommandlockouts

\usepackage{graphicx}
\usepackage{comment}
\usepackage{amsmath,amssymb} % define this before the line numbering.
\usepackage{color}
\usepackage{wrapfig}
\usepackage{multirow}
\usepackage{colortbl}
\usepackage{epsfig}
\usepackage{graphicx}
\usepackage{float}
\usepackage{subfigure}
\usepackage{xcolor}

\usepackage{graphicx}
\usepackage{amsmath}
\usepackage{amssymb}
\usepackage{booktabs}

\usepackage{xspace}
\usepackage[marginal]{footmisc}
\usepackage[pagebackref=true,breaklinks=true,colorlinks,bookmarks=false]{hyperref}
\usepackage[T1]{fontenc}

\usepackage[ruled,vlined]{algorithm2e}
\SetKwInput{KwInput}{Input}                % Set the Input
\SetKwInput{KwOutput}{Output}  
\SetKwInput{KwProcedure}{Procedure}  

\begin{document}

\title{Teacher-Student Competition for \\Unsupervised Domain Adaptation}

% author names and affiliations

\author{\IEEEauthorblockN{Ruixin Xiao$^{1}$, Zhilei Liu$^{1}$\IEEEauthorrefmark{1}\thanks{\IEEEauthorrefmark{1}Corresponding author.}, Baoyuan Wu$^{2}$}
\IEEEauthorblockA{$^{1}$ College of Intelligence and Computing, Tianjin University, Tianjin, China\\
$^{2}$School of Data Science, Chinese University of Hong Kong, Shenzhen, China\\
\{ruixinx, zhileiliu\}@tju.edu.cn, baoyuanwu1987@gmail.com\\
}
}

% make the title area
\maketitle

% As a general rule, do not put math, special symbols or citations
\begin{abstract}
With the supervision from source domain only in class-level, existing unsupervised domain adaptation (UDA) methods mainly learn the domain-invariant representations from a shared feature extractor, which causes the source-bias problem. This paper proposes an unsupervised domain adaptation approach with Teacher-Student Competition (TSC). In particular, a student network is introduced to learn the target-specific feature space, and we design a novel competition mechanism to select more credible pseudo-labels for the training of student network. We introduce a teacher network with the structure of existing conventional UDA method, and both teacher and student networks compete to provide target pseudo-labels to constrain every target sample's training in student network. Extensive experiments demonstrate that our proposed TSC framework significantly outperforms the state-of-the-art domain adaptation methods on Office-31 and ImageCLEF-DA benchmarks.
\end{abstract}

\IEEEpeerreviewmaketitle

\section{Introduction}

With the help of large-scale labeled data, supervised deep learning has achieved great progress for solving various tasks such as object detection, semantic segmentation, image classiﬁcation, etc. For many tasks, however, it is often expensive and time-consuming to collect and annotate sufficient amounts of training data. Therefore, unsupervised domain adaptation (UDA) ~\cite{gopalan2011domain} is proposed to transfer knowledge from a related source domain with rich labeled data to the unlabeled target domain. However, this UDA paradigm is hindered because of the domain shift ~\cite{torralba2011unbiased} in data distributions across domains, which is a major obstacle for transferring the model trained on the source domain to the target domain.

\begin{figure}[!ht]
  \centering
  \vspace{-2.5mm}
    \subfigure[Traditional UDA Method\label{fig:DA_mot}]{
    \begin{minipage}[]{0.45\textwidth}
    \includegraphics[width=1\textwidth]{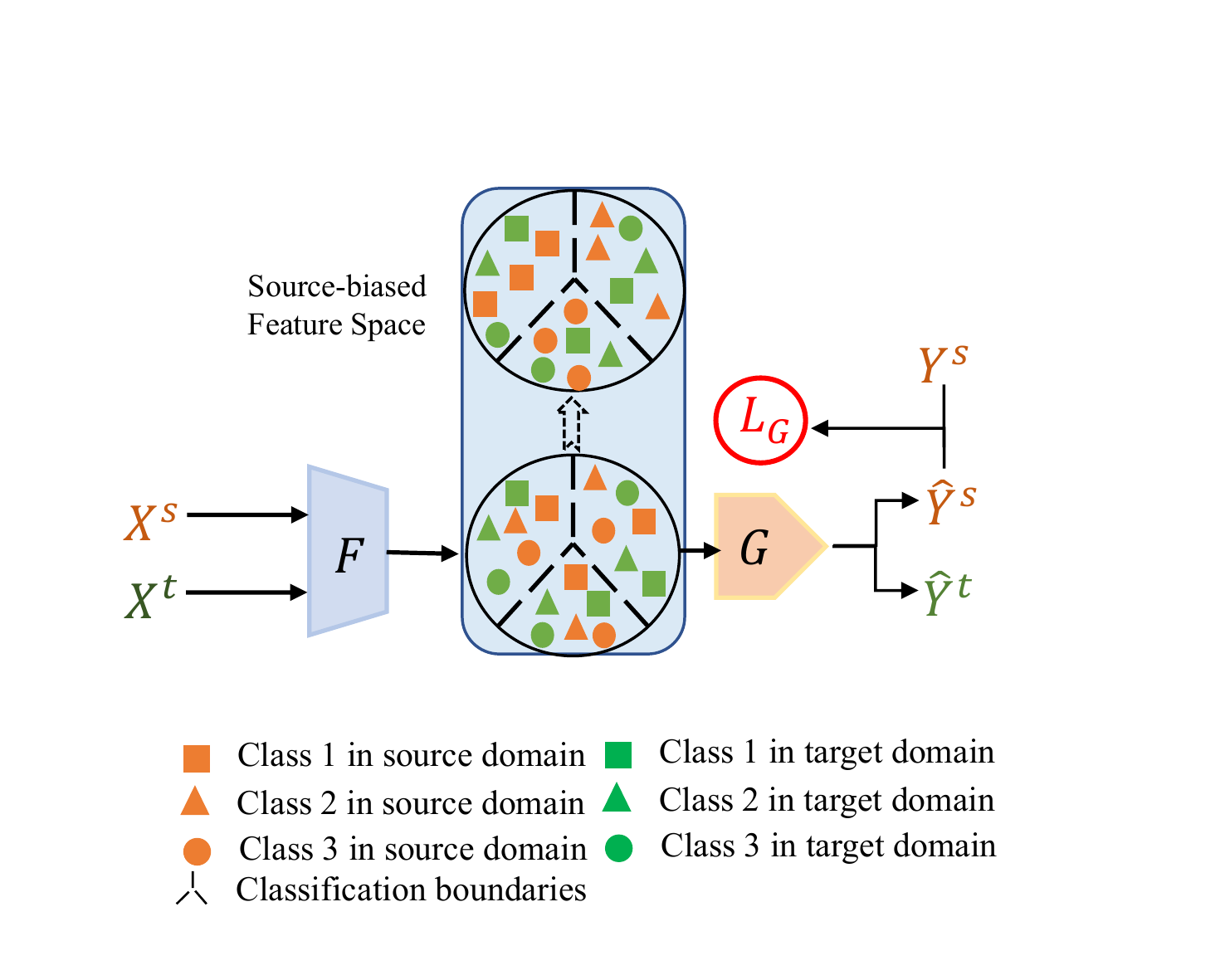}
    \end{minipage}
    }
    \subfigure[Our UDA method with TSC \label{fig:CADA_mot}]{
    \begin{minipage}[]{0.45\textwidth}
    \includegraphics[width=1\textwidth]{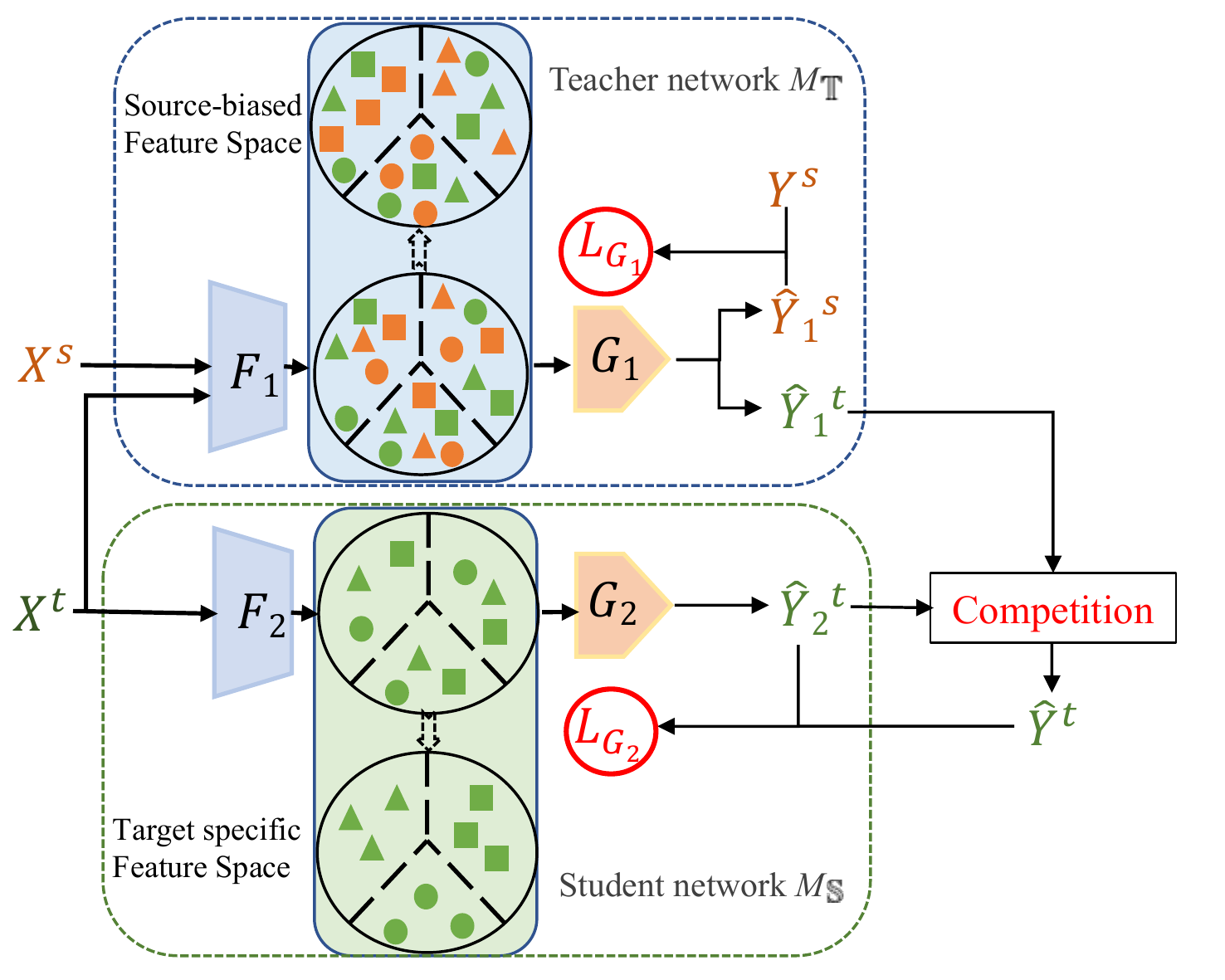}
    \end{minipage}
    }
\caption{Comparison between traditional UDA method and our method with TSC. (a) Source biased feature space is gradually learned in tradition UDA method with the supervision from source domain only. The class boundaries don't fit well for target data; (b) To alleviate the source-biased problem, our proposed TSC method aims at learning target-specific feature space in $M_\mathbb{S}$ by using pseudo-labels derived from the competition between $M_\mathbb{T}$ and $M_\mathbb{S}$. (Best viewed in color.}
	\label{fig:motivation}
	\vspace{-3mm}
\end{figure}

Mainstream UDA methods align source and target features in a common feature space to mitigate the domain shift, which can be divided into two branches. The first aims at bridging the distributions of source and target through statistics measures ~\cite{borgwardt2006integrating, long2015learning, long2017deep, yan2017mind,  zhuang2015supervised, ben2010theory, shen2018wasserstein, lee2018minimax, lee2019sliced, zellinger2017central, kang2019contrastive}, and the other methods are based on adversarial learning to extract domain-invariant feature representations, which have been reported with remarkable performances~\cite{ganin2015unsupervised, tzeng2017adversarial, long2018conditional, sankaranarayanan2018generate, chen2019transferability, pei2018multi, cao2018partial} and become a more and more significant branch in UDA. Despite their efﬁcacy, existing  methods still face critical limitations. As illustrated in Fig.~\ref{fig:DA_mot}, these methods aim at aligning feature representations extracted from a shared feature extractor $F$ to a common feature space. However, the adapted feature space is source-biased since only source data $X^s$ are annotated and the labels of target samples are unknown during training. So target feature distributions are aligned forcibly under such a source-biased criterion, which would harm their semantic information and make the classification boundaries not fit well for target domain.

To tackle the source-bias problem, this paper proposes a Teacher-Student Competition (TSC) approach for UDA. As illustrated in Fig.~\ref{fig:CADA_mot}, we build a target-specific student network $M_\mathbb{S}$, which is trained by target data only, to learn a target-specific feature space. Due to the fact that target data have no ground-truth labels during training, we constrain $M_\mathbb{S}$ by virtue of pseudo-labels. A teacher 
network $M_\mathbb{T}$ using existing conventional UDA method is further introduced to provide target predictions as a reference. The pseudo-labels $\hat{Y}_1^t$ from $M_\mathbb{T}$, however, are not used to train $M_\mathbb{S}$ directly because they are predicted based on the shared source-biased feature space and have limited effect to alleviate the source-bias problem. Instead, we design a competition module to select the more reliable pseudo-labels $\hat{Y}^t$, which is critical to break through the inherent dilemma of source-bias in $M_\mathbb{T}$ and reach a target-specific feature space in $M_\mathbb{S}$. Both $M_\mathbb{T}$ and $M_\mathbb{S}$ compete to  constrain $M_\mathbb{S}$ training. $M_\mathbb{S}$ is a simple classifier with the same structure as $M_\mathbb{T}$. The entire framework is end-to-end without any pre-processing or post-processing, and all the modules are optimized jointly.

Our motivation of competition module comes from teacher-student learning process in human learning, which is a process of chasing and surpassing. At the initial stage of learning, a student learns knowledge with the guidance of his teacher mainly. When student gets further understanding about the task, he needs to solve and makes his own predictions with higher and higher confidence. When student makes prediction with higher credibility than which of teacher in the later phase of the learning, he does believe himself even if his prediction is inconsistent with which of teacher, which is key to 
enable student to breakthrough the inherent error of teacher.

To summarize, our main contributions are threefold:
\begin{itemize}
\item To alleviate the source-biased problem, we build a target-specific network to learn a target-specific feature space.

\item We propose a novel pseudo-label selection strategy by designing a competition mechanism, in which pseudo-labels from teacher network and student network compete to be the final pseudo-label training for student network.

\item Extensive experimental results demonstrate that our method achieves state of the art performance on common benchmark domain adaptation tasks ~\cite{Imageclef-da, saenko2010adapting}. 

\end{itemize}

\section{Related Work}

We review previous works that are most relevant to our
method, including unsupervised domain adaptation, target-specific network and  pseudo-label selection strategies.

\subsection{Unsupervised Domain adaptation}

Unsupervised domain adaptation~\cite{gopalan2011domain, torralba2011unbiased}, which aims to learn a classifier on a target domain according to the knowledge transferred from a related source domain, has been developed in recent years on image classification tasks ~\cite{Imageclef-da, saenko2010adapting}. Mainstream methods on UDA can be divided into two branches. The first is to bridge the distributions between source and target domains through many different statistics measures such as maximum mean discrepancy (MMD) ~\cite{borgwardt2006integrating, long2015learning, long2017deep, yan2017mind}, kullback-lerbler (KL) divergence  ~\cite{zhuang2015supervised}, H-divergence ~\cite{ben2010theory}, wasserstein distance ~\cite{ shen2018wasserstein, lee2018minimax, lee2019sliced}, central moment discrepancy (CMD) ~\cite{zellinger2017central}, contrastive domain discrepancy(CDD) ~\cite{kang2019contrastive}, etc. The second is adversarial learning to extract domain-invariant feature representations. Inspired by generative adversarial network(GAN) ~\cite{goodfellow2014generative}, Ganin et al.~\cite{ganin2015unsupervised} proposed a DANN framework for adversarial domain adaptation by introducing a module named domain discriminator to align feature distributions. Since DANN, adversarial learning has become a more and more significant branch in domain adaptation and many methods ~\cite{ganin2015unsupervised, tzeng2017adversarial, long2018conditional, sankaranarayanan2018generate, chen2019transferability, pei2018multi, cao2018partial} have achieved excellent performances. 

However, these works align common feature space for source and target, which is overfitting to labeled source domain and cause source-bias. When feature distributions of target domain are enforced to be aligned in such a source-biased feature space, negative transfer will be introduced because of the damaged semantic information and incorrect predictions on the target domain. To address this source-biased problem, a target-specific student network is proposed to learn a target-specific feature space in this 
paper.

\subsection{Target-specific network}

Tzeng et al. ~\cite{tzeng2017adversarial} proposed an ADDA framework. It maps source and target feature extracted from a source-specific extractor and a target-specific  extractor separately to a common feature space. 
Saito et al. ~\cite{saito2017asymmetric} proposed an asymmetric tri-training strategy to learn discriminative representations for the target domain. It build a shared feature extractor and three asymmetric classifiers, in which a target-specific classifier is trained by pseudo-labels provided by another two classifiers only when their predictions about target sample are consistent and confident enough. In these methods, shared feature space is 
still the fundamental assumption and target-specific classifier is to  obtain target-discriminative representations.

In this paper, target-specific student network, especially target-specific feature extractor module in it, is built to learn a target-specific feature space, rather than to modify the shared source-biased feature space.

%%%%% Figure in Section 2
\begin{figure*}[htb!]
\centering
\includegraphics[width=0.78\textwidth]{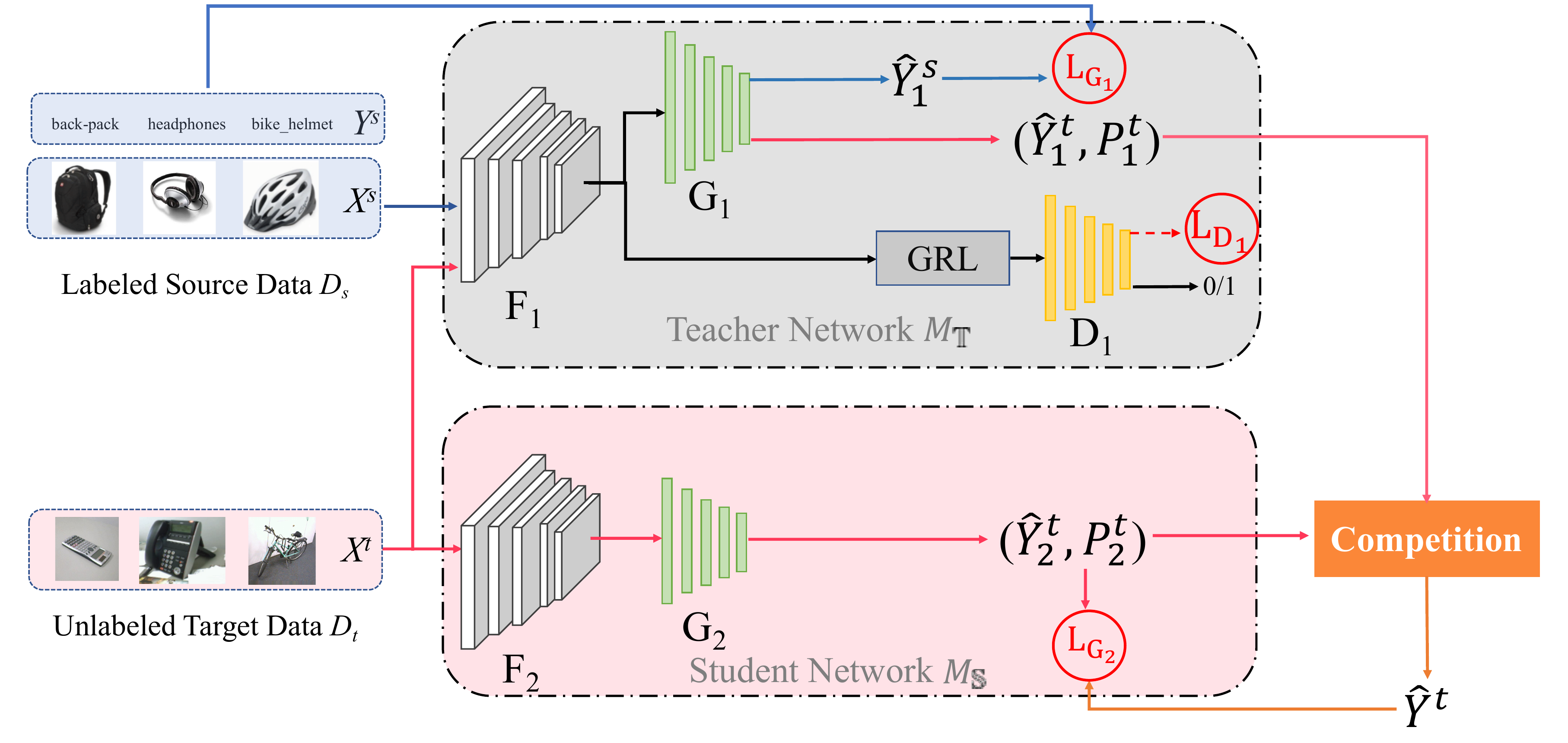}
\caption{The architecture of our TSC+DANN framework, where teacher network $M_\mathbb{T}$ (grey) and student network $M_\mathbb{S}$ (pink) feed their pseudo-labels and the corresponding probabilities  $(\hat{Y}_1^t, P_1^t)$ and $(\hat{Y}_2^t, P_2^t)$ into the competition module (orange) to determine the more credible pseudo-labels $\hat{Y}^t$ for $M_\mathbb{S}$ training. We use blue line to trace the source data and red line for target data during training, while black line in $M_\mathbb{T}$ denotes that both of the two domain data flows. Orange line from the competition module follows the trail of the selected pseudo labels. (Best viewed in color.)}
\label{fig:model}
\end{figure*}

\subsection{Pseudo-label}

To compensate the lack of categorical information in the target domain, many works ~\cite{chen2011co, saito2017asymmetric, xie2018learning, zhang2018collaborative, roy2019unsupervised, Pan_2019_CVPR, Ma_2019_CVPR, Liang_2019_CVPR, Chen_2019_CVPR, Chang_2019_CVPR} have focused on assigning pseudo-labels to target samples. One line ~\cite{xie2018learning, Pan_2019_CVPR, Ma_2019_CVPR, Liang_2019_CVPR, Chen_2019_CVPR} is to align labeled source centroids and pseudo-labeled target centroids. Another is selecting  a set of pseudo-labeled target samples as precise as possible.
%% Saito et al. ~\cite{saito2017asymmetric} proposed an asymmetric tri-training strategy to learn discriminative representations for the target domain, in which the target-specific network is trained by pseudo-labels provided by another two classifiers only when their predictions about target sample are consistent and confident enough.
 Zhang et al. ~\cite{zhang2018collaborative} iteratively selected pseudo-labeled target samples according to a threshold related to the classiﬁcation accuracy of the current image classiﬁer. Chen et al. ~\cite{Chen_2019_CVPR} developed an easy-to-hard transfer strategy (EHTS) and an adaptive prototype alignment (APA) step to train their model iteratively and alternatively with the help of a certain threshold to decide whether a pseudo-label was selected. Deng et al.~\cite{deng2019cluster} proposed cluster alignment with a teacher (CAT) based on self-ensembling~\cite{laine2016temporal, tarvainen2017mean, french2017self} to exploit the class-conditional structure in the feature space of the unlabeled target domain. Self-ensembling approaches put the stochastic transformations and perturbations in the input layers and constrains the outputs from  teacher network and student network to be consistent with each other. The teacher network in CAT is an ensemble model to provide more reliable target pseudo-labels. It is a novel approach to assign pseudo-labels for target samples, and the predictions of the pseudo-labels  are used to modify the shared feature space.

Instead of modifying the shared source-biased feature space by utilizing the pseudo-labels predicted based on this source-biased feature space, we design a teacher-student competition mechanism to select the more credible pseudo-labels $\hat{Y}^t$ for learning a target-specific student feature space. Rather than supervising $M_\mathbb{T}$ by teacher pseudo-labels singly, $M_\mathbb{T}$ and $M_\mathbb{S}$ compete to provide their pseudo-labels to train $M_\mathbb{S}$ , which is the key to enable $M_\mathbb{S}$ to break through the inherent bias in $M_\mathbb{T}$.

\section{Teacher-Student Competition for Unsupervised Domain Adaptation}

\subsection{Overview}

The general UDA model is implemented on the labeled source domain $D_s = \left \{ (x_i^s, y_i^s)  \right \}_{i=1}^{n_s}$ and the unlabeled target domain $D_t = \left \{ x_j^t  \right \}_{j=1}^{n_t}$, where $n_s$ and $n_t$ are the source and target sample numbers. The overall architecture of our proposed TSC framework is depicted in Fig.~\ref{fig:model}, which consists of three components, teacher network $M_\mathbb{T}$, student network $M_\mathbb{S}$, and the competition module. The competition module receives pseudo-labels of target samples $(\hat{Y}_1^t, P_1^t)$ and $(\hat{Y}_2^t, P_2^t)$ from $M_\mathbb{T}$ and $M_\mathbb{S}$ separately, and sends the better pseudo-labels $\hat{Y}^t$ with higher confidence to $M_\mathbb{S}$ for training. Note that $P_s$ is the distribution of source domain while $P_t$ for target, and $P_s \neq P_t$. With the help of $M_\mathbb{T}$ and competition module, the main goal of our proposed TSC based UDA method is to get a well-trained student network $M_\mathbb{S}$ mapping from the target data space $X^t$ to the target label space $Y^t$, $M_s: X^t \rightarrow Y^t$.

\subsection{Teacher Network $M_\mathbb{T}$}

Despite of their limitations, existing UDA methods have achieved excellent performances and can be adopted as good teacher networks to guide the learning of the student network by providing their results to $M_\mathbb{S}$ as a reference. The teacher network $M_\mathbb{T}$ of our proposed TSC framework is showed in grey in Fig.~\ref{fig:model}, whose objective function can be formulated as:

\begin{equation}
\label{eq:LG1D1}
\min_{F_1, G_1} \max_{D_1} \mathbf{\textit{L}}_{G_1}(F_1, G_1)-\lambda \mathbf{\textit{L}}_{D_1}(F_1, D_1)
\end{equation}

where $\lambda$ is a hyper-parameter that sets a relative trade-off. 

With the help of gradient reversal layer (GRL) ~\cite{ganin2015unsupervised} widely used in adversarial domain adaptation, feature extraction network $F_1$, classification network $G_1$ and discriminate network $D_1$ can be optimized jointly rather than alternately. 

Apart from the training process to learn the discriminate feature space, another use of $M_\mathbb{T}$ is to guide the training of $M_\mathbb{S}$. In particular, $M_\mathbb{T}$ feeds predictions and corresponding probabilities $(\hat{Y}_1^t, P_1^t)$ of target samples into the competition module to get the more credible target pseudo-labels $\hat{Y}^t$ for the training of $M_\mathbb{S}$. For a target sample $x_j^t$, once $\hat{y}_{1,j}^{t}$ is determined based on $M_\mathbb{T}$, the knowledge used for the training of $M_\mathbb{S}$ will be transferred from $M_\mathbb{T}$ to $M_\mathbb{S}$. Note that the training of $M_\mathbb{S}$ is influenced by $M_\mathbb{T}$ due to the pseudo-labels $\hat{Y}_{1}^{t}$ provided by $M_\mathbb{T}$ while $M_\mathbb{T}$ is independent of $M_\mathbb{S}$.

In fact, our TSC can be treated as a unified UDA framework, in which any existing UDA method can be used as our teacher network $M_\mathbb{T}$ to provide target pseudo-labels for $M_\mathbb{S}$ training. In this paper, two efficient UDA methods named of DANN ~\cite{ganin2015unsupervised} and CDAN ~\cite{long2018conditional} are utilized as $M_\mathbb{T}$. The details are as follows.

\subsubsection{Domain Adversarial Neural Network (DANN)}

In DANN model, a domain discriminator $D_1$ is designed to distinguish the features of source domain from the target domain, and the feature extractor $f = F_1(x)$ is trained to get features that confuse the discriminator $D_1$. It is expected to mitigate the domain shift by playing a two-player mini-max game between $F_1$ and $D_1$. The loss function of DANN is written as:

\begin{equation}
\label{eq:LG1}
    \mathbf{\textit{L}}_{G_1}(F_1, G_1)=\frac{1}{n_s} \sum_{i=1}^{n_s}\ell(G_1(F_1(x_i^s)), y_i^s)
\end{equation}

\begin{equation}
\label{eq:LD1}
    \mathbf{\textit{L}}_{D_1}(F_1, D_1)=- \frac{1}{n_s} \sum_{i=1}^{n_s}log[D_1(f_i^s)]- \frac{1}{n_t} \sum_{j=1}^{n_t}log[1-D_1(f_j^t)]
\end{equation}

 where $\mathbf{\textit{L}}_{G_1}(F_1, G_1)$ on the source classifier $G_1$ uses cross-entropy loss $\ell$ to lower the source classification error, and $\mathbf{\textit{L}}_{D_1}(F_1, D_1)$ aims at measuring the distance between the distributions of source and target data.

\subsubsection{Conditional Domain Adversarial Network (CDAN)}
Different from DANN that matches the feature representations across domains solely, CDAN takes account of discriminate information received from the classifier predictions during alignment. It conditions domain discriminator $D_1$ on the classifier prediction $g = G_1(x)$ through the multilinear map: 

\begin{equation}
\label{eq:T}
    \mathbf{T}_\otimes (h)=f \otimes g
\end{equation}

where $f= F_1(x)$ and $h=[f, g]$. So the loss function of ${L}_{D_1}(F_1, D_1)$ in CDAN can be rewritten as: 

\begin{equation}
\label{eq:LD1_CDAN}
    \mathbf{L}_{D_1}(F_1, D_1)=- \frac{1}{n_s} \sum_{i=1}^{n_s}log[D_1(h_i^s)]- \frac{1}{n_t} \sum_{j=1}^{n_t}log[1-D_1(h_j^t)]
\end{equation}

And ${L}_{G_1}(F_1, G_1)$ in CDAN is the same with that of DANN.

\subsection{Student Network $M_\mathbb{S}$}

The student network $M_\mathbb{S}$ is a simple classifier network composed of $F_2$ and $G_2$, which have the same structures as $F_1$ and $G_1$ in $M_\mathbb{T}$. To avoid getting a source-biased feature space, only target data are fed into $M_\mathbb{S}$ for training. Therefore, $M_\mathbb{S}$ is a target-specific network. However, pseudo-labels received from $M_\mathbb{T}$ contain many incorrect ones due to the effect of source-biased feature space in $M_\mathbb{T}$. So negative transfer can't be avoided if $M_\mathbb{S}$ receives pseudo-labels from $M_\mathbb{T}$ only. In our proposed TSC, $M_\mathbb{S}$ also feeds its predictions and the corresponding probabilities $(\hat{Y}_2^t, P_2^t)$ to the competition module, which is crucial to avoid the negative transfer caused by inherent incorrect pseudo-labels from $M_\mathbb{T}$ and to further modify the target feature space in $M_\mathbb{S}$. When the final pseudo label is $\hat{y}_{2,j}^{t}$ determined by our competition module instead of $\hat{y}_{1,j}^{t}$ from $M_\mathbb{T}$, the training for $M_\mathbb{S}$ is a self-training process. The objective of $M_\mathbb{S}$ is written as:

\begin{equation}
\label{eq:LG2}
    \mathbf{\textit{L}}_{G_2}(F_2, G_2)=\frac{1}{n_t}\sum_{j=1}^{n_t}\ell(G_2(F_2(x_j^t)),~\hat{y}_j^t)
\end{equation}
where $\hat{y}_j^t$ is determined based on  ${y}_{1,j}^{t}$ and ${y}_{2,j}^{t}$  by the our competition module.

\subsection{Competition Module}
\label{sec:pseudolabel}
In our TSC framework, a competition module is introduced to get more credible pseudo labels $\hat{Y}^t$ for the training of $M_\mathbb{S}$ based on $\hat{Y}_1^{t}$ and $\hat{Y}_{2}^{t}$. Different from previous teacher-student paradigms which constrain similarity of $\hat{Y}_{1}^{t}$ and $\hat{Y}_{2}^{t}$, our model takes account of the inconsistencies between teacher and student. Moreover, the predicted result of $M_\mathbb{S}$ with higher reliability than $M_\mathbb{T}$ will be regarded as $\hat{Y}^t$, which is the key to mitigate negative transfer caused by $M_\mathbb{T}$. The output pseudo labels $\hat{Y}^t$ of the competition module can be determined based on following formula:

\begin{equation}
\label{eq:yjt}
\hat{y}_j^t=\left\{\begin{matrix}
\hat{y}_{1,j}^{t},      & \text{if}~p_{1,j}^{t}>p_{2,j}^{t} \\ 
\hat{y}_{2,j}^{t},      & \text{otherwise}     
\end{matrix}\right.
\end{equation}

%\subsubsection{Teacher First.}

However, due to the fact that both $M_\mathbb{T}$ and $M_\mathbb{S}$ are trained from scratch, the pseudo-labels from both networks are not reliable at the initial training stage. With the supervision of labeled source data, the performance of $M_\mathbb{T}$ will improve as the training goes, while $M_\mathbb{S}$ will get stuck in a very poor performance since $M_\mathbb{S}$ will have converged under the constraint of unreliable target pseudo-labels at the early training stage. This phenomenon has been validated in our initial experiments. To tackle this problem, a probability threshold $T_p$ is introduced into the competition module to endow priority of pseudo label selection to $M_\mathbb{T}$ in a probability interval [$T_p$, 1], which is formulated as:

\begin{equation}
\label{eq:Tp}
\begin{matrix}
T_p=\frac{1}{1+exp({-\delta p})}
\end{matrix}
\end{equation}
where $\delta$=$10$ and $p$ denotes the training process. As the training progresses, $p$ increase to 1 from 0 and $T_p$ approach to 1 from 0.5. With the training continues, the performance of $M_\mathbb{S}$ will get more and more reliable, which will be utilized for competition with $M_\mathbb{T}$. Thus, Eq.~\ref{eq:yjt} can be modified as:

\begin{equation}
\label{eq:new_yjt}
\hat{y}_j^t=\left\{\begin{matrix}
\hat{y}_{1,j}^{t},      &~\text{if}~p_{1,j}^{t}>T_{p}~\text{or}~p_{1,j}^{t}>p_{2,j}^{t} \\ 
\hat{y}_{2,j}^{t},      & \text{otherwise}     
\end{matrix}\right.
\end{equation}

When $\hat{y}_{1,j}^{t}$ from $M_\mathbb{T}$ is selected in the competition module, the knowledge used for $M_\mathbb{S}$ training is transferred from $M_\mathbb{T}$ to $M_\mathbb{S}$. The guidance of $M_\mathbb{T}$ is significant because $M_\mathbb{T}$ is trained with the help of labeled related domain. But the inherent source biased error can affect $M_\mathbb{S}$ negatively at the same time. So it is crucial to choose $\hat{y}_{2,j}^{t}$ as supervision when it is more reliable than $\hat{y}_{1,j}^{t}$ in later training stage, which will introduce the knowledge of target domain to tackle the source bias problem.

%%%%%%%%%%%%%%%%Table in Section 4.2  Comparison with State-of-the-Arts
\begin{table*}[]
\caption{Accuracy (\%) on  ImageCLEF-DA  for unsupervised domain adaptation (ResNet-50). }
\label{tab:rst1}
\centering
\small
\begin{tabular}{l c c c c c c c}
 \toprule
\textbf{Method}       & \textbf{I to P} & \textbf{P to I} & \textbf{I to C} & \textbf{C to I} & \textbf{C to P} & \textbf{P to C} & \textbf{Avg} \\ 
\midrule
\textbf{RestNet-50~\cite{he2016deep}}   & 74.8$\pm$0.3        & 83.9$\pm$0.1        & 91.5$\pm$0.3        & 78.0$\pm$0.2        & 65.5$\pm$0.3        & 91.2$\pm$0.3        & 80.7         \\
\textbf{DAN~\cite{long2015learning}}          & 74.5$\pm$0.4        & 82.2$\pm$0.2        & 92.8$\pm$0.2        & 86.3$\pm$0.4        & 69.2$\pm$0.4        & 89.8$\pm$0.4        & 82.5         \\
\textbf{DANN~\cite{ganin2015unsupervised}}      & 75.0$\pm$0.6        & 86.0$\pm$0.3        & 96.2$\pm$0.4        & 87.0$\pm$0.5        & 74.3$\pm$0.5        & 91.5$\pm$0.6        & 85.0         \\ 
\textbf{JAN~\cite{long2017deep}}          & 76.8$\pm$0.4        & 88.0$\pm$0.2        & 94.7$\pm$0.2        & 89.5$\pm$0.3        & 74.2$\pm$0.3        & 91.7$\pm$0.3        & 85.8         \\ 
\textbf{MADA~\cite{pei2018multi}} & 75.0$\pm$0.3        & 87.9$\pm$0.2        & 96.0$\pm$0.3        & 88.8$\pm$0.3        & 75.2$\pm$0.2        & 92.2$\pm$0.3        & 85.8         \\  
\textbf{CDAN~\cite{long2018conditional}}          & 76.7$\pm$0.3        & 90.6$\pm$0.3        & 97.0$\pm$0.4        & 90.5$\pm$0.4        & 74.5$\pm$0.3        & 93.5$\pm$0.4        & 87.1         \\ 
\textbf{CDAN+E~\cite{long2018conditional}}      & 77.7$\pm$0.3        & 90.7$\pm$0.2        & \textbf{97.7}$\pm$0.3        & 91.3$\pm$0.3        & 74.2$\pm$0.2        & 94.3$\pm$0.3        & 87.7         \\ 

\textbf{iCAN~\cite{zhang2018collaborative}} & 79.5       & 89.7        & 94.7        & 89.9        & 78.5        & 92.0        & 87.4         \\  
\textbf{rDANN+CAT~\cite{deng2019cluster}} & 77.2$\pm$0.2        & 91.0$\pm$0.3        & 95.5$\pm$0.3        & 91.3$\pm$0.3        & 75.3$\pm$0.6        & 93.6$\pm$0.5        & 87.3         \\ 
\midrule
\textbf{TSC+DANN}     &78.3$\pm$0.2       &92.8$\pm$0.3       &96.8$\pm$0.2       &90.3$\pm$0.5       &74.5$\pm$0.7       &96.0$\pm$0.2       &88.1              \\ 
\textbf{TSC+CDAN}     &\textbf{79.0}$\pm$0.3       &\textbf{93.2}$\pm$0.5       &97.2$\pm$0.4     &\textbf{92.7}$\pm$0.2       &\textbf{77.4}$\pm$0.4   &\textbf{96.5}$\pm$0.3                 &\textbf{89.3}             \\ 
\bottomrule
\end{tabular}
\end{table*}

\subsection{Overall Objective Function}

The overall objective function of our proposed TSC can be formulated as:

\begin{equation}
\label{eq:LAll}
\mathbf{\textit{L}}=\mathbf{\textit{L}}_{G_1}(F_1, G_1)-\lambda \mathbf{\textit{L}}_{D_1}(F_1, D_1)+\beta\mathbf{\textit{L}}_{G_2}(F_2, G_2)
\end{equation}

where $\lambda$ and $\beta$ are trade-off parameters for $M_\mathbb{T}$ and $M_\mathbb{S}$ separately. Algorithm~\ref{alg:CADA} shows the optimization of TSC. 

\begin{algorithm}[hbt!]
\small
\SetAlgoLined

\KwInput{
\begin{itemize}
    \item $D_s = \left \{ (x_i^s, y_i^s)  \right \}_{i=1}^{n_s}$ and $D_t = \left \{ x_j^t  \right \}_{j=1}^{n_t}$
\end{itemize}
}
\KwOutput{
\begin{itemize}
    \item Target label $\left \{ \hat{y}_{1,j}^t  \right \}_{j=1}^{n_t}$ from the optimized $F_2^*$ and $G_2^*$
\end{itemize}
}
\KwProcedure{\\
          \While {not converged}{
              \begin{itemize}
                 \item Mini-batch sampling from $D^s$ and $D^t$:\\ $(X^s,Y^s)=\left \{ (x_i^s, y_i^s)  \right \}_{i=1}^{b_s}$, $X^t=\left \{ x_j^t  \right \}_{j=1}^{b_t}$
                 
                 \item  Forward $(X^s, X^t)$ into $M_\mathbb{T}$, \\compute Eq.~\ref{eq:LG1D1} with $Y^s$ and get $\{(\hat{y}_{1,j}^t, p_{1,j}^t)\}_{j=1}^{b_t}$
                 
                 \item Forward $X^t$ into $M_\mathbb{S}$ to get $\{(\hat{y}_{2,j}^t, p_{2,j}^t)\}_{j=1}^{b_t}$
                 
                 \item Feed $\{(\hat{y}_{1,j}^t,p_{1,j}^t)\}_{j=1}^{b_t}$ and $\{(\hat{y}_{2,j}^t, p_{2,j}^t)\}_{j=1}^{b_t}$  into \\competition module to get $\{\hat{y}_{j}^t\}_{j=1}^{b_t}$
                 
                 \item Compute Eq.~\ref{eq:LG2} with $\{\hat{y}_{j}^t\}_{j=1}^{b_t}$
                 
                 \item Update $F_1$, $G_1$ and $D_1$ by Eq.~\ref{eq:LG1D1}, \\update $F_2$ and $G_2$ by Eq.~\ref{eq:LG2}
            \end{itemize}
  		  }%\endWhile
      }
\caption{Description of our proposed TSC.}
\label{alg:CADA}
\end{algorithm}

%%%%%%%%%%%%%%%%Table in Section 4.2  Comparison with State-of-the-Arts
\begin{table*}[]
\caption{ Accuracy (\%) on  Office-31  for unsupervised domain adaptation (ResNet-50). }
\label{tab:rst2}
\centering
\small
\begin{tabular}{l c c c c c c c}
 \toprule
\textbf{Method}       & \textbf{A to W} & \textbf{D to W} & \textbf{W to D} & \textbf{A to D} & \textbf{D to A} & \textbf{W to A} & \textbf{Avg} \\ 
\midrule
\textbf{RestNet-50~\cite{he2016deep}}   & 68.4$\pm$0.2        & 96.7$\pm$0.1        & 99.3$\pm$0.1        & 68.9$\pm$0.2        & 62.5$\pm$0.3        & 60.7$\pm$0.3        & 76.1         \\
\textbf{DAN~\cite{long2015learning}}          & 80.5$\pm$0.4        & 97.1$\pm$0.2        & 99.6$\pm$0.1        & 78.6$\pm$0.2        & 63.6$\pm$0.3        & 62.8$\pm$0.2        & 80.4         \\
\textbf{DANN~\cite{ganin2015unsupervised}}      & 82.0$\pm$0.4        & 96.9$\pm$0.2        & 99.1$\pm$0.1        & 79.7$\pm$0.4        & 68.2$\pm$0.4        & 67.4$\pm$0.5        & 82.2         \\ 
\textbf{JAN~\cite{long2017deep}}          & 85.4$\pm$0.3        & 97.4$\pm$0.2        & 99.8$\pm$0.2        & 84.7$\pm$0.3        & 68.6$\pm$0.3        & 70.0$\pm$0.4        & 84.3         \\ 
\textbf{MADA~\cite{pei2018multi}} & 75.0$\pm$0.3        & 87.9$\pm$0.2        & 96.0$\pm$0.3        & 88.8$\pm$0.3        & 75.2$\pm$0.2        & 92.2$\pm$0.3        & 85.8         \\  
\textbf{GTA~\cite{sankaranarayanan2018generate}}          & 89.5$\pm$0.5        & 97.9$\pm$0.3        & 99.8$\pm$0.4        & 87.7$\pm$0.5        & 72.8$\pm$0.3        & 71.4$\pm$0.4        & 86.5         \\ 
\textbf{CDAN~\cite{long2018conditional}}          & 93.1$\pm$0.2        & 98.2$\pm$0.2        & \textbf{100.0}$\pm$0.0        & 89.8$\pm$0.3        & 70.1$\pm$0.4        & 68.0$\pm$0.4        & 86.6         \\ 
\textbf{CDAN+E~\cite{long2018conditional}}      & 94.1$\pm$0.1        & 98.6$\pm$0.1        & \textbf{100.0}$\pm$0.0        & 92.9$\pm$0.2        & 71.0$\pm$0.3        & 69.3$\pm$0.3        & 87.7         \\ 
\textbf{iCAN~\cite{zhang2018collaborative}} & 92.5       & \textbf{98.8}        & \textbf{100.0}        & 90.1        & 72.1       & 68.9        & 87.2         \\  
\textbf{rDANN+CAT~\cite{deng2019cluster}} & 94.4$\pm$0.1        & 98.0$\pm$0.2        &\textbf{ 100.0}$\pm$0.0       & 90.8$\pm$1.8        & 72.2$\pm$0.6        & 70.2$\pm$0.1        & 87.6         \\  
\midrule
\textbf{TSC+DANN}     &85.0$\pm$0.3    &98.0$\pm$0.1     &\textbf{100.0}$\pm$0.0      &80.3$\pm$0.4      & 69.3$\pm$0.2       & 67.7$\pm$0.1        & 83.4     \\ 

\textbf{TSC+CDAN}     &\textbf{94.6}$\pm$0.3     &98.2$\pm$0.2     &\textbf{100.0}$\pm$0.0      & \textbf{94.7}$\pm$0.1      & \textbf{74.0}$\pm$0.2    & \textbf{71.6}$\pm$0.7      &\textbf{88.9}     \\ 
\bottomrule
\end{tabular}
\end{table*}

\section{Experiments}
\subsection{Datasets and Settings}
\label{sec:setting}

\subsubsection{Datasets}

Our TSC framework is evaluated on two benchmarks for UDA task: ImageCLEF-DA ~\cite{Imageclef-da} and Office-31~\cite{saenko2010adapting}.

\begin{itemize}

    \item \textbf{ImageCLEF-DA}  is a benchmark dataset originally used for the ImageCLEF 2014 domain adaptation challenge. It contains three domains: \textit{ImageNet ILSVRC 2012} (\textbf{I}), \textit{Pascal VOC 2012} (\textbf{P}), and \textit{Caltech-256} (\textbf{C}). Each domain has 12 categories and contains 50 images per category. We conduct experimental evaluations on 6 transfer tasks. 
    
    \item \textbf{Office-31} is a large scaled dataset for domain adaptation, which contains 4,652 images in 31 categories from 3 domains: \textit{Amazon} (\textbf{A}), \textit{Webcam} (\textbf{W}) and \textit{DSLR} (\textbf{D}). We consider 6 transfer tasks by using all domain combinations during experimental evaluation.
    
\end{itemize}

\subsubsection{Implementation Details}

We follow the standard protocols for unsupervised domain adaptation ~\cite{ganin2015unsupervised, long2017deep} by using all labeled source examples and all unlabeled target examples at the training stage. We compare the average classification accuracy based on three random experiments. The pre-trained ResNet-50~\cite{he2016deep} on ImageNet~\cite{russakovsky2015imagenet} are further fine-tuned as the feature extractor in our TSC. Our TSC is trained with Stochastic Gradient Descent (SGD) using PyTorch, a mini-batch size of 36, a weight decay of 0.0005, and a momentum of 0.95. The learning rate of the layers trained from scratch are set to be 10 times those of ﬁne-tuned layers. We ﬁx $\lambda=1.0$ and $\beta=0.3$ in all experiments.

\subsection{Comparison with State-of-the-Arts}

We extend our proposed TSC with DANN~\cite{ganin2015unsupervised} and CDAN~\cite{long2018conditional}, named as TSC+DANN and TSC+CDAN, respectively. We compare our proposed TSC method  against state-of-the-art deep learning based domain adaptation methods under the same setting stated in Section~\ref{sec:setting}: ResNet-50 ~\cite{he2016deep}, Deep Adaptation Network (DAN) ~\cite{long2015learning}, Domain Adversarial Neural Network (DANN) ~\cite{ganin2015unsupervised}, Joint Adaptation Network (JAN)  ~\cite{long2017deep}, Generate to Adapt (GTA) ~\cite{sankaranarayanan2018generate}, MADA ~\cite{pei2018multi} Conditional Domain Adversarial Network (CDAN) ~\cite{long2018conditional}, Incremental Collaborative and Adversarial Network (iCAN)   ~\cite{zhang2018collaborative} and Cluster Alignment with a Teacher(CAT) ~\cite{deng2019cluster}. For a fair comparison, the results of all state-of-art methods are either directly reported from their original papers wherever available or quoted from ~\cite{long2018conditional}.

\textbf{Evaluation on ImageCLEF-DA}. The results on the ImageCLEF-DA dataset are reported in Table~\ref{tab:rst1}.  It can be observed that our proposed TSC methods outperform state of the art results on most DA tasks and boost the performance of DANN and CDAN significantly. In particular, our TSC+DANN and TSC+CDAN bring relative increments of $3.1\%$ and $2.2\%$ for average accuracy over DANN and CDAN, respectively. The promotion on DANN is more significant than CDAN, which is reasonable since CDAN takes class information into account for aligning feature space. Moreover, we can observe that our TSC method is highly dependent on $M_\mathbb{T}$. The higher performance $M_\mathbb{T}$ has, the higher performance $M_\mathbb{S}$ gets.

\begin{figure}[htb!]
\centering
\includegraphics[width=0.45\textwidth]{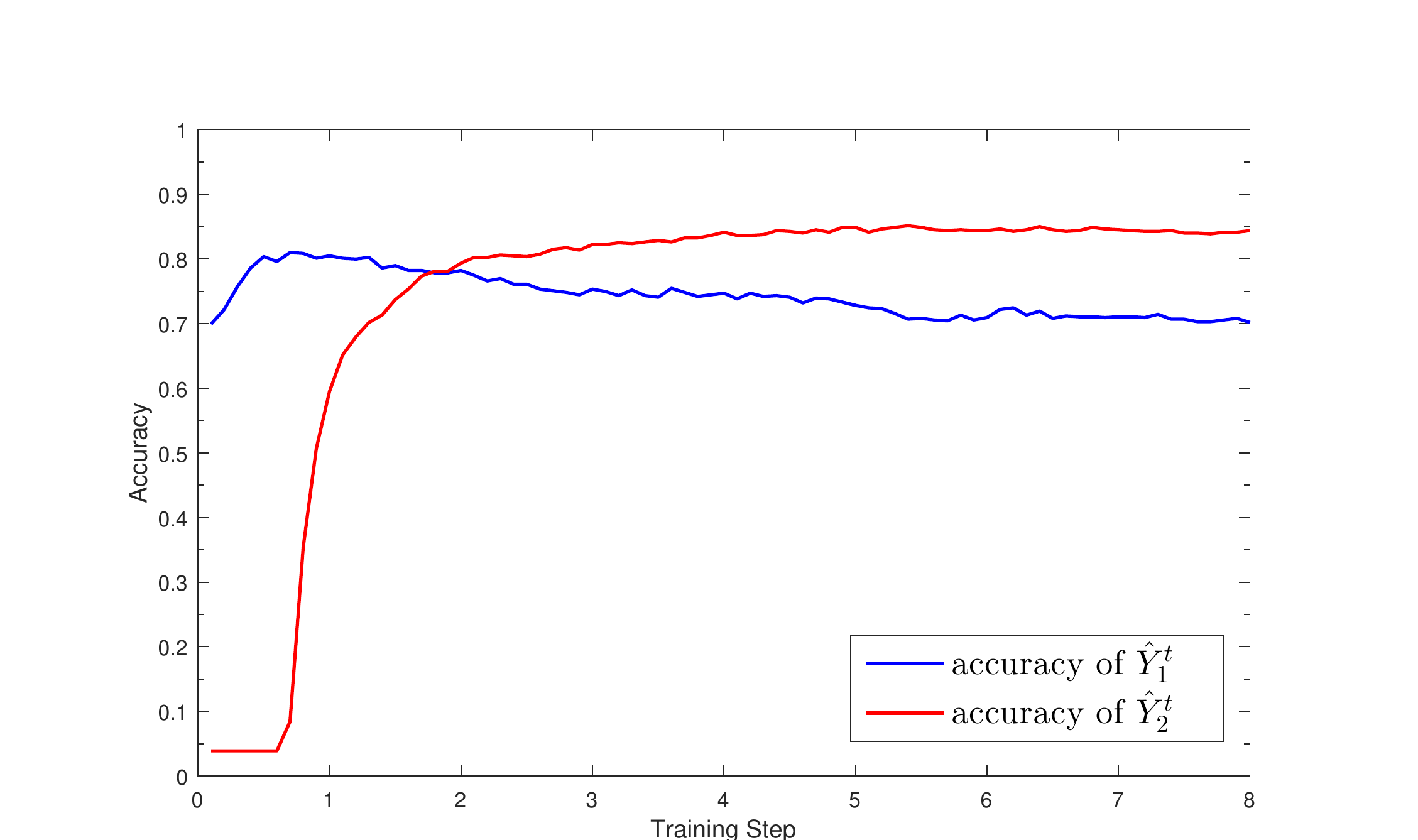}
\caption{Accuracy of TSC+CDAN on \textbf{A}$\rightarrow$\textbf{W} in office-31}
\label{fig:CADA_dann_a2w}
\end{figure}

\textbf{Evaluation on Office-31}. The accuracy results evaluated on Office-31 are shown in Table~\ref{tab:rst2}. It can be seen that our TSC framework with CDAN as $M_\mathbb{T}$ significantly outperforms all the previous works on most of DA tasks. In particular, our TSC+DANN and TSC+CDAN bring relative increments of $1.2\%$ and $2.3\%$ for average accuracy over DANN and CDAN, respectively. Note that, even though our TSC+DANN outperforms DANN significantly, we have poor performances on the tasks \textbf{A}$\rightarrow$\textbf{W}, \textbf{A}$\rightarrow$\textbf{D}, \textbf{D}$\rightarrow$\textbf{A}, and \textbf{W}$\rightarrow$\textbf{A} comparing with the most recent state-of-the-art methods. The reason is that it is difficult to reach a good and steady performance for teacher network $M_\mathbb{T}$ using DANN. Taking task \textbf{A}$\rightarrow$\textbf{W} as an example in Fig.~\ref{fig:CADA_dann_a2w}, we can observe that the accuracy of $M_\mathbb{T}$ (DANN) drops significantly after it reaches its peak at the early stage. With a lower performance of $M_\mathbb{T}$ at the later stage, it is harder for our competition module to choose credible pseudo-labels for the training of $M_\mathbb{S}$, since both $M_\mathbb{T}$ and $M_\mathbb{S}$ present poor performances.

\begin{figure}[t]
  \centering
  \vspace{-2.5mm}
    \subfigure[TSC+DANN ($M_\mathbb{T}$) \label{fig:Pseudo_a}]{
    \begin{minipage}[]{0.45\textwidth}
    \includegraphics[width=1\textwidth]{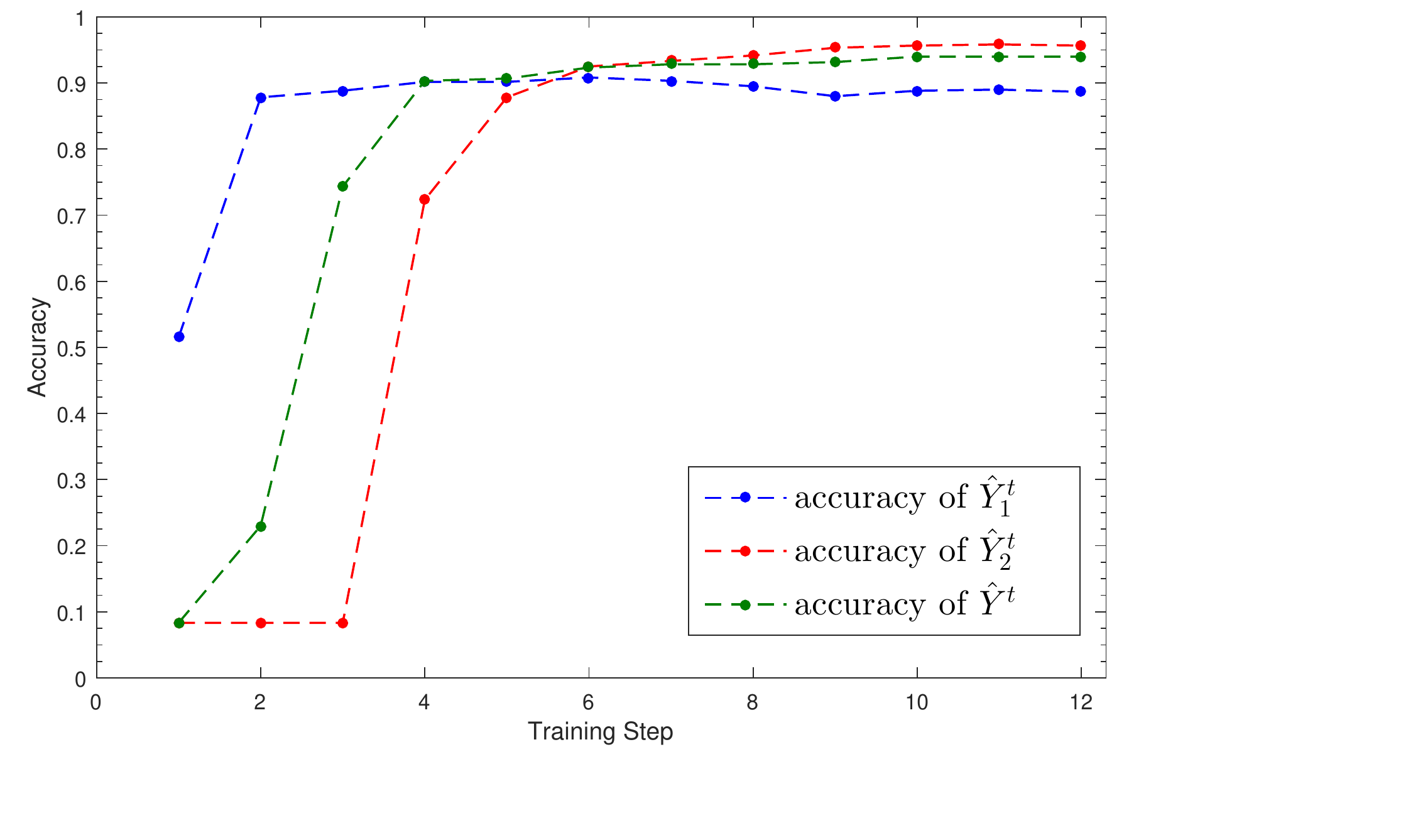}
    \end{minipage}
    }
    \subfigure[TSC+CDAN ($M_\mathbb{T}$) \label{fig:Pseudo_b}]{
    \begin{minipage}[]{0.45\textwidth}
    \includegraphics[width=1\textwidth]{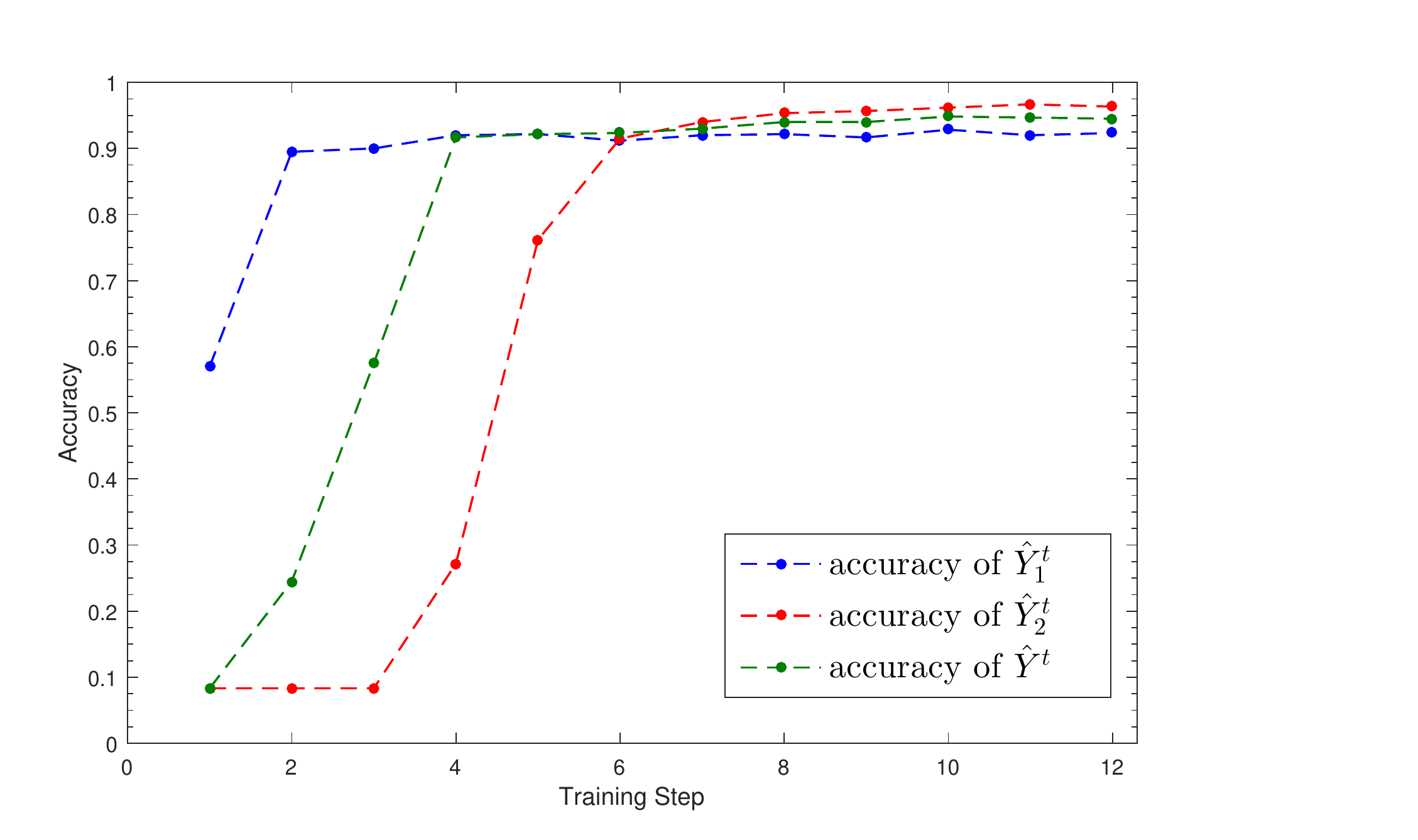}
    \end{minipage}
    }
\caption{Pseudo-labeling accuracy comparison on task \textbf{P}$\rightarrow$\textbf{C} in ImageCLEF-DA. (a) $M_\mathbb{T}$ is DANN; (b) $M_\mathbb{T}$ is CDAN. }
    \label{fig:Pseudo_Acc}
	\vspace{-3mm}
\end{figure}

\begin{figure*}[t]\vspace{-0.05in}
  \centering
  \subfigure[$M_\mathbb{T}$ in TSC+DANN \label{fig:Vis_a}]{
  \begin{minipage}[t]{0.22\textwidth}
   \centering
    \includegraphics[width=1\textwidth]{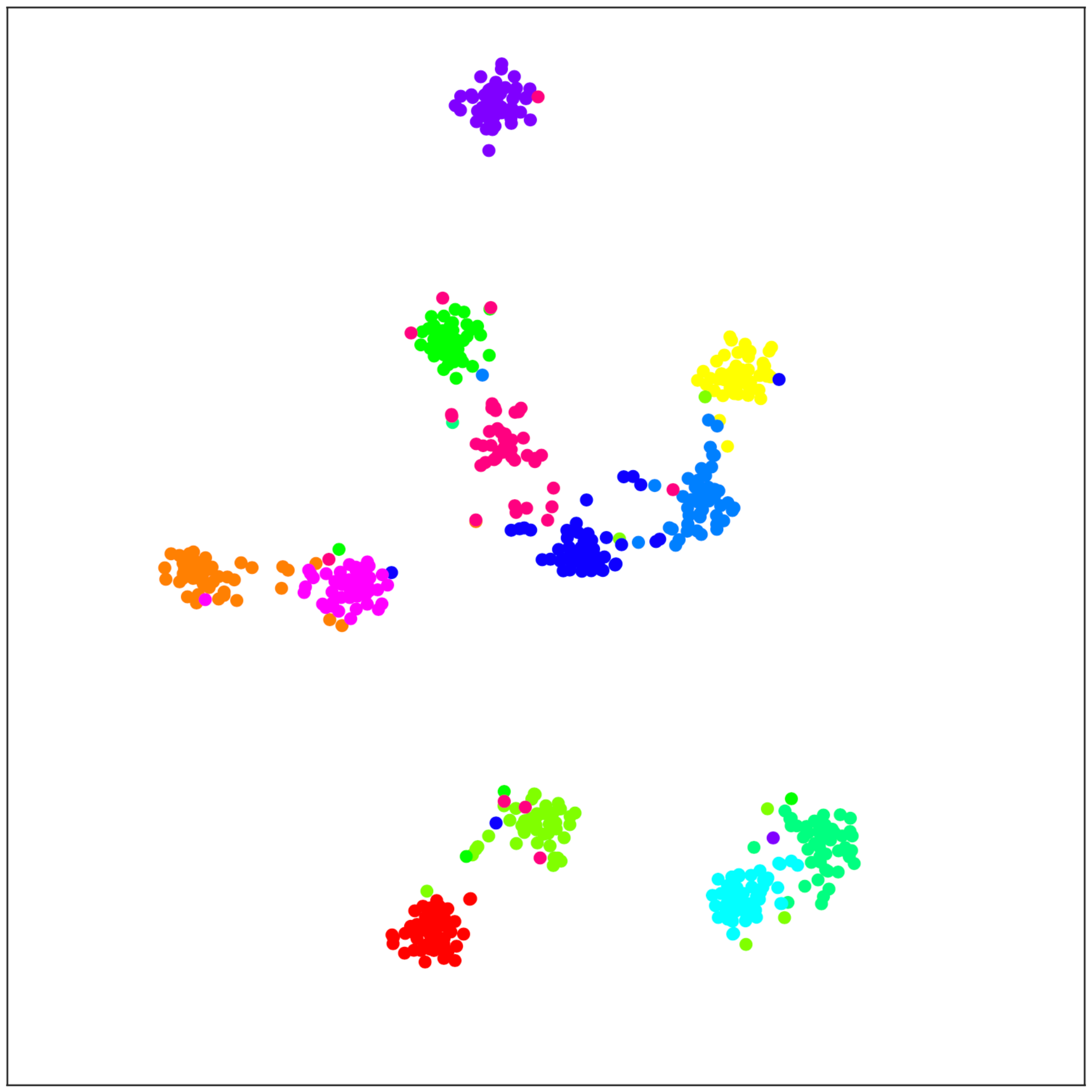}
    \end{minipage}
  }
  \subfigure[$M_\mathbb{S}$ in TSC+DANN \label{fig:Vis_b}]{
  \begin{minipage}[t]{0.22\textwidth}
   \centering
    \includegraphics[width=1\textwidth]{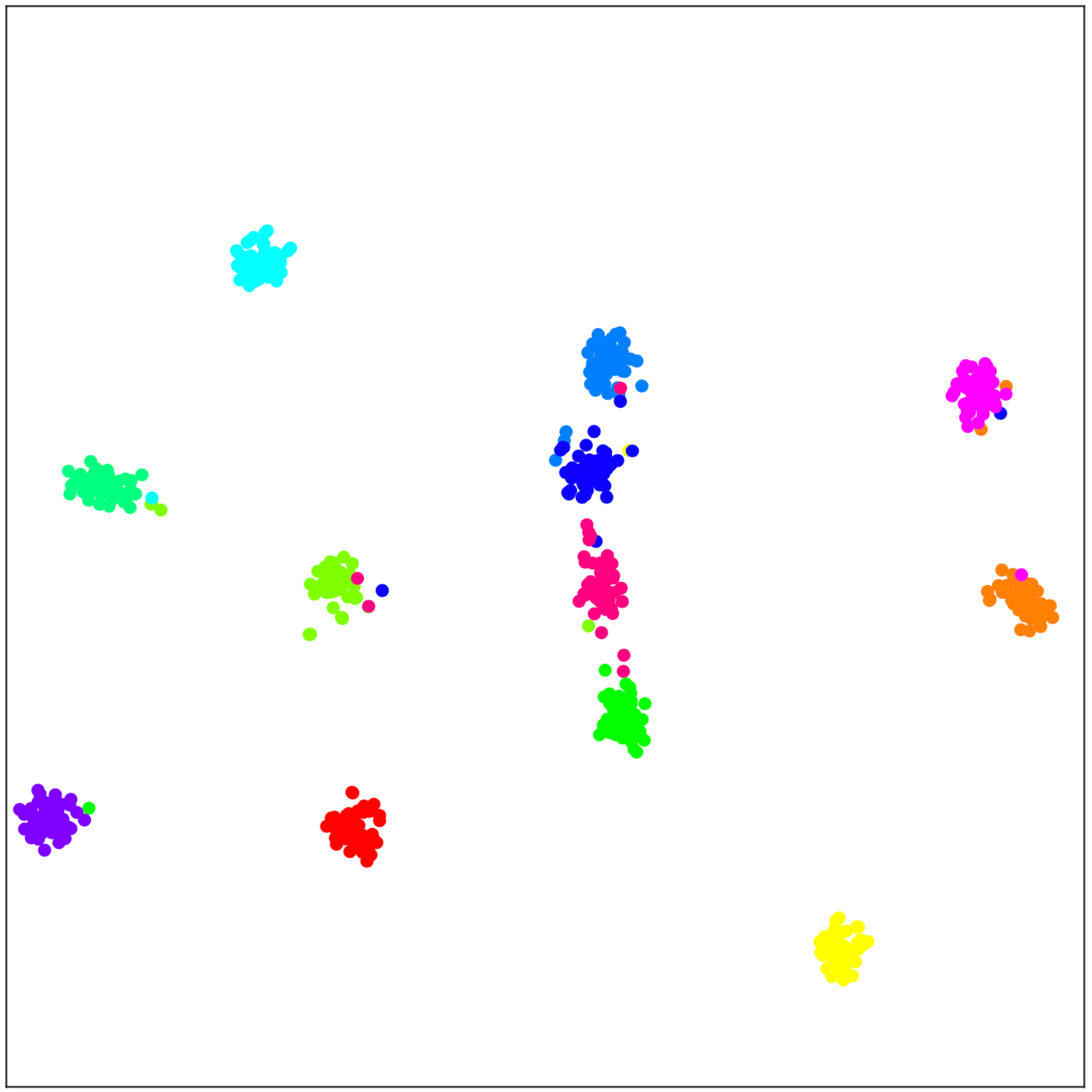}
    \end{minipage}
    }
  \subfigure[$M_\mathbb{T}$ in TSC+CDAN  \label{fig:Vis_c}]{
  \begin{minipage}[t]{0.22\textwidth}
   \centering
    \includegraphics[width=1\textwidth]{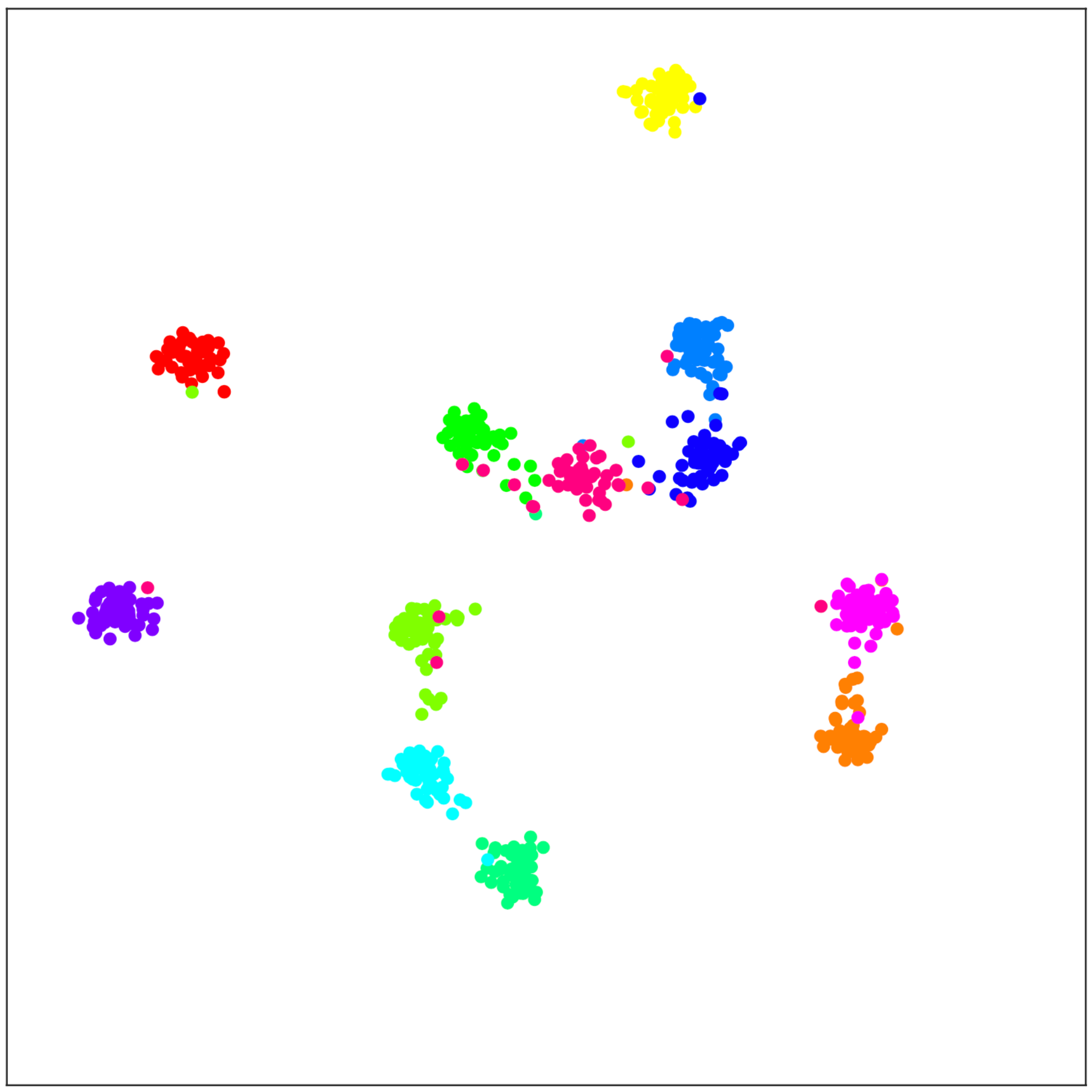}
    \end{minipage}
  }
    \subfigure[$M_\mathbb{S}$ in TSC+CDAN  \label{fig:Vis_d}]{
  \begin{minipage}[t]{0.22\textwidth}
   \centering
    \includegraphics[width=1\textwidth]{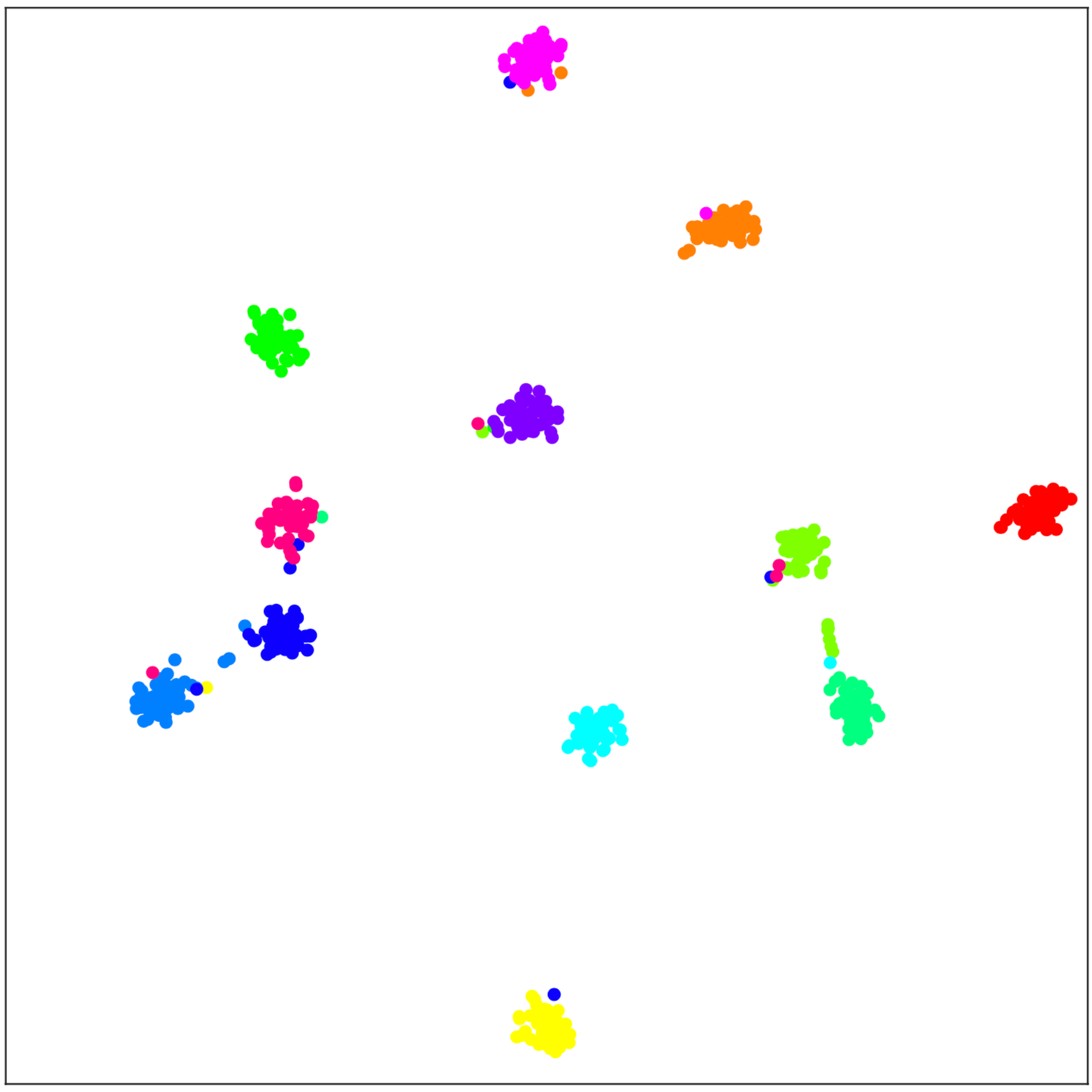}
    \end{minipage}
  }
% Requires \usepackage{graphicx}
%   \includegraphics[width=0.45\textwidth]{fig/wa_weight_distribution.pdf}
 \caption{The t-SNE visualization of target features on task \textbf{P}$\rightarrow$\textbf{C} in ImageCLEF-DA. Different colors represent different categories (best viewed in color). (a) $M_\mathbb{T}$ is DANN in TSC+DANN. (b) $M_\mathbb{S}$ in TSC+DANN. (c) $M_\mathbb{T}$ is CDAN in TSC+CDAN. (d) $M_\mathbb{S}$ in TSC+CDAN. }
    \label{fig:Vis}
\end{figure*}

\subsection{Pseudo-labeling Accuracy}

To verify the effectiveness of our competition module during training, Fig.~\ref{fig:Pseudo_Acc} illustrates three kinds of pseudo-label accuracy curve, i.e. $\hat{Y}_{1}^{t}$ from $M_\mathbb{T}$ , $\hat{Y}_{2}^{t}$ from $M_\mathbb{S}$, and $\hat{Y}^{t}$ from the competition module. Since the training of $M_\mathbb{T}$ is independent of $M_\mathbb{S}$, so blue curves in Fig.~\ref{fig:Pseudo_a} and Fig.~\ref{fig:Pseudo_b} show the performance fluctuations of DANN and CDAN as the training goes.

At the early stage of the training process, we can observe that $M_\mathbb{S}$ tends to converge to a single category with low accuracy. The reason is that most prediction probabilities of $M_\mathbb{T}$ are too low  to surmount the threshold $T_p$, so priority to $M_\mathbb{T}$ works weakly when selecting the winning pseudo-labels. Therefore, under the main rule of fair competition, the ignorant $M_\mathbb{S}$ dominate the competition module, utilizing almost all of its own target predictions to supervise its training so early that it gets stuck with poor performance. As the training goes on, the priority to $M_\mathbb{T}$ is activated when the  prediction probabilities of $M_\mathbb{T}$ increase over $T_p$. So the accuracy of winning pseudo-label goes up with the hoist of $M_\mathbb{T}$, helping $M_\mathbb{S}$ out of the trap of convergent low accuracy. As the training goes further, the strategy of priority to $M_\mathbb{T}$ dies away as $T_p$ approach to 1. It is shown that the accuracy curve of winning pseudo-label surpasses $M_\mathbb{T}$. And $M_\mathbb{S}$ also surpasses $M_\mathbb{T}$ with the guidance of winning pseudo-label when $M_\mathbb{S}$ trains. We can see that the competition mechanism does select more accurate pseudo-labels than $M_\mathbb{T}$ itself to constrain the learning process of $M_\mathbb{S}$.

At the later stage of learning, the performance of $M_\mathbb{S}$ climbs to top and exceeds to the winning pseudo-label line. On the beneficial aspect, it means that the pseudo-labels generated by our target-specific classifier $M_\mathbb{S}$ are more accurate than not only $M_\mathbb{T}$ but also winning pseudo-labels from competition mechanism even if some of them fail in the competition. On the bad aspect, we train $M_\mathbb{S}$ with the supervision of winning pseudo-labels that has lower accuracy than $M_\mathbb{S}$ itself, which preserves $M_\mathbb{S}$ from improving performance further. How to break this dilemma and improve performance further is the challenge in our further work.

\subsection{Feature Visualization}

Taking task \textbf{P}$\rightarrow$\textbf{C} (12 classes) as an example in Fig.~\ref{fig:Vis}, we utilize t-SNE ~\cite{maaten2008visualizing} to visualize the target feature representations learned by $M_\mathbb{T}$ and its corresponding $M_\mathbb{S}$. Note that, the visualized results of $M_\mathbb{T}$ and $M_\mathbb{S}$ in TSC+DANN are shown in Fig.~\ref{fig:Vis_a} and Fig.~\ref{fig:Vis_b} respectively, and the visualized results of $M_\mathbb{T}$ and $M_\mathbb{S}$ in TSC+CDAN are shown in Fig.~\ref{fig:Vis_c} and Fig.~\ref{fig:Vis_d} respectively. No matter when $M_\mathbb{T}$ is DANN or CDAN, we can observe that a) Our $M_\mathbb{S}$ learns more discriminative representations, while some categories in $M_\mathbb{T}$ have been mixed up in the feature space. b) The 12 clusters formed by the features generated from $M_\mathbb{S}$ are more compact and have more clear inter-class boundaries. c) $M_\mathbb{S}$ has less  classified flasely samples compared to $M_\mathbb{T}$. All of these show that our $M_\mathbb{S}$ can learn more discriminative target feature representations and modify the source-biased feature space in $M_\mathbb{T}$ to get a more unbiased feature space in $M_\mathbb{S}$.

\section{Conclusion}

To alleviate the source biased problem in conventional UDA methods, this paper presents unsupervised domain adaptation approach with Teacher-Student competition (TSC). It introduces a student network to learn a target-specific feature space and a teacher network with the structure of existing conventional UDA method to provide target pseudo-labels as reference.  To select more credible pseudo-labels for the training of student network, a novel competition mechanism is designed. Both teacher and student network compete to provide target pseudo-labels to constrain every target sample's training in student network. Extensive experiments implemented on Office-31 and ImageCLEF-DA benchmarks demonstrate that (i) our proposed TSC significantly outperforms the state-of-the-art domain adaptation methods, and (ii) more separable target feature space can be achieved by introducing our competition model to tackle the source-bias problem. However, we set the threshold $T_p$ as a hyper parameter manually and experimentally in this paper, which is not flexible enough and limits the performance to improve further. In the future work, we consider to design it dynamically and smoothly.

% conference papers do not normally have an appendix

% use section* for acknowledgment
\section*{Acknowledgment}
This work is supported by the National Natural Science Foundation of China (No. 61503277) and the Tianjin Municipal Science and Technology Project (No. 19ZXZNGX00030). It is also partially supported by the joint research project with China Automotive Technology and Research Center.

\bibliographystyle{IEEEtran}
\bibliography{reference}

% that's all folks
\end{document}